\documentclass{article}

\usepackage{microtype}
\usepackage{graphicx}
\usepackage{subcaption}
\usepackage{booktabs} 

\usepackage{multirow}   

\usepackage{hyperref}
\usepackage{xurl}

\usepackage[preprint]{icml2026}

\usepackage{enumerate, amsmath, amsthm, amssymb, pifont, csquotes, dashrule, tikz, bbm, booktabs, bm, verbatim, url, mathrsfs, mathtools}
\usepackage[framemethod=TikZ]{mdframed}
\usepackage{natbib}

\usepackage[capitalize,noabbrev]{cleveref}

\theoremstyle{plain}

\theoremstyle{definition}

\theoremstyle{remark}

\usepackage[textsize=tiny]{todonotes}

\newif\ifsubmit
\submittrue
\ifsubmit
\newcommand{\dnote}[1]{}
\newcommand{\tnote}[1]{}
\newcommand{\rnote}[1]{}
\newcommand{\cnote}[1]{}
\newcommand{\wnote}[1]{}
\else
\newcommand{\dnote}[1]{\textcolor{blue}{Dilip: #1}}
\newcommand{\tnote}[1]{\textcolor{orange}{Tom: #1}}
\newcommand{\rnote}[1]{\textcolor{purple}{Ryan: #1}}
\newcommand{\cnote}[1]{\textcolor{teal}{Ced: #1}}
\newcommand{\wnote}[1]{\textcolor{green}{Wil: #1}}
\fi

\icmltitlerunning{Cognitive Models and AI Algorithms Provide Templates for Designing Language Agents}

\definecolor{dblue}{RGB}{98, 140, 190}
\definecolor{dlblue}{RGB}{216, 235, 255}
\definecolor{dgreen}{RGB}{124, 155, 127}
\definecolor{dpink}{RGB}{207, 166, 208}
\definecolor{dyellow}{RGB}{255, 248, 199}
\definecolor{dgray}{RGB}{46, 49, 49}

\newcommand{\durl}[1]{\textcolor{dblue}{\underline{\url{#1}}}}

\newcommand{\mc}[1]{\mathcal{#1}}

\newcommand{\bN}{\mathbb{N}}

\newcommand{\ra}{\rightarrow}

\DeclareMathOperator*{\argmax}{arg\,max}

\newmdenv[
  topline=false,
  bottomline=false,
  rightline = false,
  leftmargin=10pt,
  rightmargin=0pt,
  innertopmargin=0pt,
  innerbottommargin=0pt
]{innerproof}

\newcounter{DaveDefCounter}
\begin{document}

\twocolumn[
  \icmltitle{Cognitive Models and AI Algorithms Provide Templates \\ for Designing Language Agents}



  \icmlsetsymbol{equal}{*}

  \begin{icmlauthorlist}
    \icmlauthor{Ryan Liu}{equal,pcs}
    \icmlauthor{Dilip Arumugam}{equal,pcs}
    \icmlauthor{Cedegao E. Zhang}{equal,mit} \\
    \icmlauthor{Sean Escola}{columbia}
    \icmlauthor{Xaq Pitkow}{cmu}
    \icmlauthor{Thomas L. Griffiths}{pcs,ppsy}
  \end{icmlauthorlist}

  \icmlaffiliation{pcs}{Department of Computer Science, Princeton University}
  \icmlaffiliation{mit}{Department of Brain and Cognitive Sciences, Massachusetts Institute of Technology}
  \icmlaffiliation{columbia}{Zuckerman Mind Brain Behavior Institute and Department of Psychiatry, Columbia University}
  \icmlaffiliation{cmu}{Neuroscience Institute and Department of Machine Learning, Carnegie Mellon University}
  \icmlaffiliation{ppsy}{Department of Psychology, Princeton University}

  \icmlcorrespondingauthor{Ryan Liu}{\texttt{ryanliu@princeton.edu}}
  \icmlcorrespondingauthor{Dilip Arumugam}{\texttt{dilip.a@cs.princeton.edu}}
  \icmlcorrespondingauthor{Ced Zhang}{\texttt{cedzhang@mit.edu}}

  \icmlkeywords{Machine Learning, ICML}
  \vskip 0.3in
]



\printAffiliationsAndNotice{}  

\begin{abstract}
  While contemporary large language models (LLMs) are increasingly capable in isolation, there are still many difficult problems that lie beyond the abilities of a single LLM. For such tasks, there is still uncertainty about how best to take many LLMs as parts and combine them into a greater whole. This position paper argues that potential blueprints for designing such modular language agents can be found in the existing literature on cognitive models and artificial intelligence (AI) algorithms. To make this point clear, we formalize the idea of an agent template that specifies roles for individual LLMs and how their functionalities should be composed. We then survey a variety of existing language agents in the literature and highlight their underlying templates derived directly from cognitive models or AI algorithms. By highlighting these designs, we aim to call attention to agent templates inspired by cognitive science and AI as a powerful tool for developing effective, interpretable language agents.
\end{abstract}

\section{Introduction}

Recent research in artificial intelligence (AI) has increasingly focused on creating language agents---systems based on large language models (LLMs) interacting with one another or computational tools to more effectively perform a set of tasks~\citep{su-etal-2024-language, wang2024survey, liu2025advances, xi2025rise}. A challenge in this area is identifying effective agent designs---specifications of the roles LLMs should play and how they should interact~\citep{sumers2023cognitive, cemri2025multi}. The large search space of possible agent architectures makes brute force exploration infeasible and successful designs can often appear to be arbitrary. While prolonged iteration over possible designs may work in some cases, practitioners trying to solve real-world problems in high-stakes settings seldom have the data and resources to do so, making it important to develop effective heuristics for designing language agents. 

Here, we argue for the position that \textbf{cognitive models and existing AI algorithms provide templates for designing language agents}. Cognitive models and AI algorithms have been informing each other since the beginning of their respective fields \cite{newell1958elements, minsky1986society}. Here, we show that they provide effective design templates for language agents; the architecture of any language agent consists of distinct processes that are carefully ordered and executed sequentially to solve a problem. Many applications of interest for language agents involve problems that are addressed by existing cognitive models and AI algorithms that are explicitly expressed in terms of such modular, sequential processing. Moreover, recent work has produced software tools that support the modular design and deployment of such language agents at scale~\citep{vezhnevets2023generative}.

To support our position, we show how existing agent architectures can be seen as instantiating cognitive models or AI algorithms, with applications that range from supporting effective communication \cite{liu2023improving} to improving planning \cite{webb2025brain} to exploring efficiently \cite{arumugam2025toward}. 
While cognitive models are typically defined in narrow domains matching the assumptions of specific behavioral experiments, a cognitive model identifies processes that can be implemented with LLMs and indicates how those processes should interact. Likewise, AI algorithms pick out procedural solutions to problems such as searching through a set of hypotheses or gathering information to select actions. 
Using LLMs to implement language agents based on cognitive models or AI algorithms broadens the applicability of existing models and algorithms with the flexibility and expressivity of natural language, making them applicable in new settings and at larger scales.

To make this argument, we first provide an explicit definition of \textbf{agent templates} (Section~\ref{sec:templates}). We define an agent template as a specification of a set of functions and the interactions between those functions. 
We then give concrete examples of how cognitive models (Section \ref{sec:cognitive}) and AI algorithms (Section \ref{sec:algorithms}) have previously been used as agent templates. With these templates, the goals of an agent can be transformed into a pipeline defining how exactly the agent should break down and complete the task. This approach increases human interpretability and provides a coherent and previously-successful strategy for addressing many problems faced by language agents. 

\section{Background}

\subsection{Language agents}

Language agents are LLM-based autonomous agents that can follow language instructions to carry out diverse and complex tasks in real-world or simulated environments~\citep{su-etal-2024-language, wang2024survey}. They have been used for various tasks including coding~\citep[e.g.,][]{wang2024openhands, cognition_2024_devin, anthropic_claude_3_7_sonnet_2025, yang2024swe}, customer service chatbots~\citep{fin2024ai, Sierra2026, Decagon2026}, research~\citep{openai_deep_research_2025, sakana_ai_scientist_2024, lu2024ai}, and grounded robotics tasks~\citep[e.g.,][]{ahn2022can, driess2023palm, raptis2025agentic}. 
Agent designs often augment LLMs with goals and planning~\citep{yao2023react, park2023generative}, persistent state or memory~\citep{shinn2024reflexion, packer2023memgpt, wang2025agent}, actions/tool use~\citep{schick2023toolformer, wang2024tools, patil2024gorilla}, and autonomy in order to complete tasks over multiple LLM calls~\citep{wang2023voyager, wu2024autogen}.
Most LLM agents leverage an instruction-tuned foundation model and construct agent scaffolding via prompting~\citep{karpas2022mrkl, sahoo2024systematic, luo2025large}, although other approaches exist that further use fine-tuning~\citep{chen2023fireact, xu2023rewoo} or RL~\citep{yao2022webshop, zhai2024fine, qi2024webrl, li2025torl}. 

Since language agents often tackle more complicated tasks~\citep{wang2024survey} and contain a multitude of components (e.g., memory) that each require implementation-level choices~\citep{zhang2025survey, sumers2023cognitive}, the number of potential architectures to search through rises accordingly. In reality, practitioners rarely search through the entirety of this space, instead relying on folk intuition, manual design, or even opportunistic chance to build their design~\citep{zamfirescu2023johnny, wang2024survey, cui2025automatic}. This choice is often partially due to the limited data available to conduct a search over architectures in domains in which AI agents are deployed~\citep{liu2025survey, amodei2016concrete, kaufmann2023survey, fu2020batch, dulac2019challenges} including settings such as healthcare~\citep{tang2021model}. While a common method to address this is to simulate human data~\citep[e.g.,][]{wang2024mint, yao2024tau}, such solutions may not represent an adequate distribution of human rewards~\citep{seshadri2026lost, yang2024regularizing, wang2024llm}. 

\subsection{Agent design frameworks}

The definition of agent templates we introduce in the next section is not the first attempt to formalize modular LLM systems. \citet{zhang2025agentic} introduce agentic context engineering as an agent-design framework consisting of a generator, reflector, and curator; while the generator processes queries with access to a dynamic ``playbook,'' a reflector evaluates the generator's successes and shortcomings so that they may be summarized by the curator for integration into the next playbook. In parallel, \citet{he2025information} introduce compressor-predictor systems as an agent-design framework consisting of a compressor LLM that compactly summarizes input data into a succinct context that may be leveraged by the subsequent predictor to produce outputs. While both frameworks encapsulate a number of agent designs, they operate at a strictly lower level of abstraction than our proposed agent templates. More importantly, these high-level decompositions are somewhat arbitrary and lack a degree of flexibility toward any problem. 

\citet{sumers2023cognitive} share our high-level goal of connecting language agent research with cognitive science. They introduce the CoALA framework inspired by symbolic cognitive architectures research and use it to organize a wide body of work on agents, categorized in terms of relevant concepts (e.g., memory) and actions (e.g., retrieval). However, they leave the question of \textit{how} to build such agents for future work and do not address concrete connections to either cognitive models or AI algorithms. Our work fills in these gaps and highlights the value of these connections.

Closest to our agent templates are compound AI systems~\citep{khattab2023dspy, compound-ai-blog, agrawal2025gepa} and  GPTSwarm~\citep{zhuge2024gptswarm}. Both frameworks also propose a directed graph structure over LLMs and tools for composing language agents. Crucially, however, \citet{khattab2023dspy} and \citet{zhuge2024gptswarm} both approach the corresponding agent design problem given either a compound AI system or GPTSwarm as an optimization problem. While they demonstrate successful results on a variety of tasks with either a genetic algorithm for evolving an agent design or explicit reinforcement learning, such approaches fails to recognize the plethora of existing solutions for numerous sub-problems in cognitive science and AI. Our core position in this paper is that acknowledging these existing solution concepts and embedding them into an agent design not only circumvents a laborious optimization process but also lends a tremendous degree of interpretability to performant language agents.

\section{Defining templates for language agents}
\label{sec:templates}

To provide common structure to our analysis of cognitive models and AI algorithms, this section provides a formal definition of a template for a language agent.

For any arbitrary set $A$, let $\mc{P}(A)$ denote the power set of $A$. For any two arbitrary sets $\mc{X}$ and $\mc{Y}$, let $\{\mc{X} \ra \mc{Y}\} \triangleq \{f \mid f:\mc{X} \ra \mc{Y}\}$ denote the class of all functions mapping inputs in $\mc{X}$ to outputs in $\mc{Y}$.

Let $V$ be a finite vocabulary of tokens, $|V| < \infty$, and let $L \in \bN$ denote a maximum length. Language is defined as the space of all possible token sequences consisting of no more than $L$ tokens, $\mc{L} \triangleq \bigcup\limits_{n=1}^L V^n$. Abstractly, any language model is a stochastic function $f:\mc{L} \ra \Delta(\mc{L})$ mapping input language (a prompt) to a distribution over possible output language, which is then sampled to yield a response. More broadly, modern language agents are often endowed with tools that can provide dedicated access to privileged information (e.g., search or database query APIs) or functionality (e.g., document and image processing APIs); regardless, however, the inputs to such tools originate in language provided by the agent and the corresponding outputs must ultimately be textualized before being returned. Thus, it suffices to let $\mc{F} \triangleq \{\mc{L} \ra \Delta(\mc{L})\}$ denote the set of all possible modules for designing a language agent.

We formalize an agent template as a directed, acyclic graph (DAG) $\mc{G} = (\mc{V}, \mc{E})$ where vertices $\mc{V} \subseteq \mc{F}$ are either LLMs or tools and edges $\mc{E} = \mc{V} \times \mc{V}$ denote connections between these vertices. The parents of any vertex $v \in \mc{V}$ in $\mc{G}$ can be defined by a function $\Gamma: \mc{V} \ra \mc{P}(\mc{V})$ which yields all other vertices in $\mc{G}$ associated with an edge directed to $v$, $\Gamma(v) \triangleq \{v' \in \mc{V} \mid (v',v) \in \mc{E}\}$. The initial or root vertices of $\mc{G}$ are defined as those exclusively appearing in edges where they are the tail or source, $\mc{R} \triangleq \{v \in \mc{V} \mid \nexists (v', v) \in \mc{E}, \forall v' \in \mc{V} \} \subset \mc{V}$. Conversely, the terminal vertices of $\mc{G}$ are those only appearing in edges where they are the head or destination, $\mc{T} \triangleq \{v \in \mc{V} \mid \nexists (v, v') \in \mc{E}, \forall v' \in \mc{V} \} \subset \mc{V}$.

Intuitively, an individual agent template specifies a collection of one or more LLMs and tools as well as the execution order and flow of data between those modules. For any input $\ell \in \mc{L}$, initial vertices $r \in \mc{R}$ could be executed in parallel right away while any subsequent module $v \in \mc{V} \setminus \mc{R}$ could only be called once the outputs of all its parents in $\Gamma(v)$ are available as input. In order to guarantee a sensible execution path, we assume that $\mc{G}$ is not only acyclic but also weakly connected such that, by treating all edges in $\mc{E}$ as undirected, any pair of vertices in $\mc{V}$ are connected by at least one path. For ease of exposition, we always think of an agent template $\mc{G}$ as consuming a single input $\ell \in \mc{L}$; when needed, we will accommodate multiple language inputs by interpreting them as being concatenated together to form the single $\ell$, which is then passed through $\mc{G}$ and can be decomposed by the constituent LLMs as needed. Similarly, the overall number of outputs for an agent template can be determined by the number of terminal vertices, $|\mc{T}|$. As LLMs are straightforward input-output mappings, this infrastructure for handling multiple inputs or outputs may seem unnecessary. Importantly, however, the subsequent examples we present of templates derived from cognitive model and AI algorithms will sometimes require consuming, updating, and emitting additional stateful information or metadata. For example, an agent template attempting to emulate count-based exploration for RL agents would need to consume additional language outlining existing count information, potentially update the counts based on recent visitation, and then output the updated count information for use by the language agent in the next time period. 

In contrast to prior works on modular DAG formalisms~\citep{thomas2011policy,weber2019credit,chang2019automatically, chang2021modularity}, an agent template is simply a vehicle for expressing possible language agent designs. We now show that inspiration for such templates can come from existing cognitive models (Section \ref{sec:cognitive}) and AI algorithms (Section \ref{sec:algorithms}). 

\section{Templates based on cognitive models}
\label{sec:cognitive}

Cognitive models are used by cognitive scientists to model how parts of the mind, including the brain, works. Each cognitive model represents a mechanistic hypothesis grounded in cognitive processes. They have the same inputs (e.g., psychological stimuli) and outputs (e.g., choices, reaction times) as humans, and are tested against collected human data. In each section below we highlight how cognitive models have been used to design agents in specific domains. 

\subsection{Communication}
Linguistic communication has been long studied in cognitive science~\citep{posner1989foundations, bermudez2014cognitive, noveck2018experimental}. A key effort has been to develop cognitive models that capture communicative choices that people make~\citep{desaussure1916, shannon1948, grice1957meaning,  grice1975logic, lewis1969convention, stalnaker1978}. One of the most prominent such models is the Rational Speech Acts (RSA)~\citep{frank2012, goodman2016pragmatic, degen2023rational}, which views communication as a rational choice between multiple utterances with corresponding utilities~\citep{simon1955behavioral}. %
Within this framework, listeners and speakers engage in recursive social inference, reasoning about each other's beliefs in order to decide what to say. A common method of such reasoning is to imagine others' reactions to each utterance choice~\citep[episodic future thinking;][]{atance2001episodic}. This capacity to project oneself into the future to pre-experience an event is closely linked to memory and past experiences~\citep{szpunar2010episodic, schacter2007remembering, klein2010facing}.

One paper that builds upon these ideas for LLMs is \citet{liu2023improving}, where the agent's goal is to recommend an utterance for a user-provided communicative scenario. To do this, the agent first creates a list of advice, then generates utterance candidates based on combinations of advice, builds profiles of key audience groups, and finally simulates audience reactions to candidates to find the best option, 
corresponding to the template $\mc{V} = \{\Lambda_{\mathrm{advisor}}, \Lambda_{\mathrm{generator}}, \Lambda_{\mathrm{profiler}}, $ $\Lambda_{\mathrm{simulator}}, \Lambda_{\mathrm{aggregator}}\}$. 
In this process, episodic future thinking is made explicit by simulating audiences' reactions. 
This entire process falls within a single step of recursive social inference outlined in RSA, and outperforms baselines and ablations according to human judgments --- representing an adequate agent implementation of a cognitive model.

Similarly, \citet{qiu2025wavelength} applies RSA to improve the communication of LLMs in the reference game Wavelength.\footnote{\url{wikipedia.org/wiki/Wavelength_(game)}} The system generates a set of possible utterances given context from the LLM, uses separate LLM calls to evaluate the likelihood of context given each utterance, and samples the final utterance in a Bayes-rational way---also explicitly implementing mental simulation. This approach significantly improved the performance of a wide range of LLMs for both direct generation and chain-of-thought~\citep{wei2022chain}. 

Other papers have also borrowed ideas around recursive social inference. 
\citet{kim2025hypothesis} follow a Bayesian theory-of-mind approach, using LLMs to approximate probabilistic inference over agents' mental states based on their percepts and actions.
And \citet{zhang2025k} investigated LLMs’ hierarchical social reasoning (i.e., higher order beliefs) following classic psychological literature~\citep{perner1985john, hedden2002you, goodie2012levels}.

\subsection{Reasoning and planning}

Reasoning is one of the oldest topics in the study of mind, going back at least to Aristotle \citep{aristotle1984prior}. Cognitive science has explored many different aspects of reasoning \citep{holyoak2005cambridge, kahneman2011thinking, tenenbaum2011how, evans2013rationality}, including intuitive theories \citep{gerstenberg2017intuitive, ullman2017mind}, judgments \citep{tversky1974judgment, gilovich2002heuristics}, problem solving \citep{newell1972human, halpern2013thought}, and counterfactuals \citep{byrne2005rational, gerstenberg2024counterfactual}. A key methodological tool for studying human reasoning is the verbal protocol or ``think aloud'' method~\citep{ericsson1993protocol, van1994think}, in which participants verbalize their thought processes. 
Reasoning models like OpenAI o1 \citep{openai2024o1} and DeepSeek R1 \citep{guo2025deepseek} trained with long chain of thought by reinforcement learning, which follows the simple $\mc{V} = \{\Lambda_{\mathrm{reasoning}}\}$ template, may be viewed as a particular manifestation of the principles behind such verbal protocols. Recent work has begun to explore how human reasoning traces can inform and improve LLM reasoning (and vice versa), especially given that LLMs themselves can help scale human verbal data collection \citep{xie2024evaluating,wurgaft2025scaling, de2025cost}. For example, \citet{kargupta2025cognitive} provide a categorization of elements of reasoning grounded in cognitive science literature and identifies gaps and shortcomings in LLMs' reasoning traces. 

Planning is a cornerstone capability for intelligent agents, without which one cannot accomplish complex goals \citep{russell2020artificial}. In cognitive science, planning is typically viewed as a kind of reasoning---reasoning about \textit{what to do} \citep{miller1960plans, bratman1987intention}. \citet{webb2025brain} show how an agent template based on cognitive modeling can improve planning with language agents. Their proposed Modular Agentic Planner (MAP) has six modules, which are responsible for task decomposition, action proposal, error monitoring, state prediction, and task coordination and correspond to the template $\mc{V} = \{\Lambda_{\mathrm{decomposer}}, \Lambda_{\mathrm{actor}}, \Lambda_{\mathrm{monitor}}, \Lambda_{\mathrm{predictor}}, \Lambda_{\mathrm{evaluator}},$ $\Lambda_{\mathrm{orchestrator}}\}$. Each module, implemented by an LLM, takes inspirations from how the prefrontal cortex is believed to function---a brain region generally involved in decision making and planning. The combined planner achieves strong results on multi-step problem solving benchmarks such as Tower of Hanoi and PlanBench. Additionally, MAP has a tree search component for action selection, drawing on a large body of research has used approximate search algorithms to model human planning \citep{ho2022people, mattar2022planning, van2023expertise, collins2025people}. We explicitly discuss agent templates derived from search algorithms in Sec.~\ref{sec:search} below.

\subsection{Representation}

Cognitive science has used symbolic representations to describe human mental processes \citep{boole1854investigation, fodor1975language, johnson1983mental, newell1994unified}, although how such representations are implemented is still debated \citep{lake2017building, santoro2021symbolic, griffiths2025whither}. The modern interpretation of the ``Language of Thought'' (LoT) hypothesis posits that many aspects of thinking and learning can be modeled as writing and executing code in some general-purpose programming language \citep{goodman2014concepts, rule2020the, quilty2023best}. In other words, the mind might contain operations like variable manipulation, conditional branching, and recursions, and can leverage appropriate data structures and algorithms in response to task demands. Programs provide more versatile and flexible representations of knowledge and skills compared to other formalisms like logical formulas and graphical models \citep{probmods2, chollet2019measure, griffiths2024bayesian}. This approach has been used to characterize a wide range of human cognitive behaviors, such as concept learning \citep{goodman2008rational, piantadosi2011learning, yang2022one} and commonsense reasoning \citep{wong2023word, zhang2023grounded, ying2025language, wong2025modeling}. 

Language agents have demonstrated the power of programming languages as a general representational device. Most agent frameworks using these techniques can be described by the template $\mc{V} = \{\Lambda_{\mathrm{reasoner}}, \Lambda_{\mathrm{interpreter}}\}$, where $\Lambda_{\mathrm{reasoner}}$ is an LLM that generates code (possibly along with natural language), and $\Lambda_{\mathrm{interpreter}}$ is a system that executes the code. Some of the first works that instructs LLMs to write programs to reason include PaL \citep{gao2023pal} and Program of Thoughts \citep{chen2023program}: instead of having the LLM carry out reasoning in natural language, a code snippet is generated and the execution result becomes the answer. These approaches provide significant performance gains on mathematical, financial, and symbolic reasoning tasks. The CodeAct agent design \citep{wang2024executable} demonstrates that for tasks where an LLM agent needs to interact with some external environment (e.g., calling API tools or navigating embodied environments), using programs to represent actions delivers performance gains over alternative text or JSON representations \citep[cf.][]{yao2023react}. Chain of Code \citep{li2023chain} proposes that LLM-generated code does not need to be fully defined to be effective---functions like detecting sarcasm can be emulated by the same LLM (in this case the template includes another $\Lambda_{\mathrm{emulator}}$). More recently, the CodeAdapt framework \citep{zhang2025code} shows that instruct LLMs plus multi-turn code use can outperform their reasoning-trained counterparts (e.g., DeepSeek V3 vs.~R1) over a diverse range of tasks such as instruction following and creativity tests, while increasing token efficiency. These results illustrate that program representations can be just as effective for designing language agents as they are for describing aspects of human cognition.

\section{Templates from AI Algorithms}
\label{sec:algorithms}

Existing AI algorithms provide tested solutions to many of the problems faced by language agents, and hence a source of agent templates. We break down our analysis into two categories---classic AI algorithms, and more recent work based on reinforcement learning (RL) algorithms. 

\subsection{Templates from Classic Algorithms}

\subsubsection{Search}
\label{sec:search}

The historical ``good, old-fashioned AI`` period in artificial intelligence research was characterized by the representation of real-world entities as symbols and the use of graphical or tree search to construct intelligent systems~\citep{mccarthy1955proposal,newell1956logic}. 
A search algorithm identifies a strategy for traversing a graph (where the edges correspond to actions or search operators and the nodes correspond to intermediate states) to identify a solution. Breadth-first search~\citep{moore1959shortest} and depth-first search~\citep{tarjan1972depth} naturally proceed with breadth-first and depth-first traversals of the graph, respectively. Extending Dijkstra's Algorithm~\citep{dijkstra1959note}, $A^*$ search~\citep{hart1968formal} identifies a minimum cost path between specified start and end states while using an admissible heuristic to prioritize which (partial) paths are explored first. Beam search~\citep{lowerre1976harpy,rubin1977locus} extends best-first search~\citep{pohl1970heuristic,pearl1985heuristics}, which greedily looks for a minimum-cost path according to a heuristic and improves memory efficiency by only retaining the top-ranked candidates in its so-called ``beam'' after each expansion. 

Agent templates derived from search algorithms implement and orchestrate components for node expansion and evaluation as language modules in order to solve a variety of search problems. For LLM inference, a standard approach for boosting accuracy on supervised-learning tasks is chain-of-thought prompting. One interpretation of this question-reasoning-answering process is that each individual question induces a graph where nodes represent partial computations or solutions. In this light, standard chain-of-thought prompting can be seen as greedy best-first search with respect to some unknown, latent heuristic codified into the underlying LLM over the course of pre-training and fine-tuning. 

An alternative choice is to externalize this heuristic search process and make it explicit in the agent's choices for which partial solution to expand further and how to value the resulting modifications. This exact process was formalized through the Tree of Thoughts framework~\citep{yao2023tree}, which concretely leverages two distinct LLMs for generating possible reasoning extensions from a current partial solution as well as evaluating the quality of progression towards a solution from an incomplete search state: $\mc{V} = \{\Lambda_{\mathrm{generator}}, \Lambda_{\mathrm{evaluator}}\}$. Here, one may interpret the graphical structure given by the agent template as a representation of input-output processing during a single iteration of the search algorithm. \citet{yao2023tree} presented and evaluated two separate agent templates, varying the underlying classic search algorithm --- breadth-first search or depth-first search --- used to govern the logic around which LLM is called on what search state. In  a similar attempt to enhance the reasoning capabilities of language agents, \citet{xie2023self} leverage an agent design inspired by (stochastic) beam search, where the two generation and evaluation LLMs are further complemented by a third ``correctness'' LLM whose resulting assessment of partial reasoning step's correctness is then used to determine priority within the beam search that iteratively prunes less-promising reasoning chains from consideration. Moving beyond the reasoning capabilities problem to general exploration of multi-modal agents within web interfaces, \citet{koh2025tree} offer an instantiation of Tree of Thoughts utilizing the $A^*$ search algorithm and with novel application of multi-modal models (for processing both language and visual features of websites) for state expansion and evaluation.

We conclude this section by mentioning a final search algorithm of increasing popularity and ubiquity: Monte-Carlo Tree Search (MCTS)~\citep{coulom2006efficient,browne2012survey}. Each iteration of MCTS consists of four steps: (1) selection of a node for further search using a designated choice of tree policy, (2) expansion of the selected node by selecting one more actions from the particular state, (3) simulation from a newly expanded state to a terminal state, and (4) backpropagation of value information to all incident states between the path from root to terminal state. Modeling the selection and expansion phases as a multi-armed bandit problem~\citep{lattimore2020bandit} and taking inspiration from upper-confidence methods for provably-efficient bandit exploration~\citep{auer2002using}, a widely adopted choice of tree policy is Upper Confidence Bound for Trees (UCT)~\citep{kocsis2006bandit} that selects based on mean reward estimates along with a carefully-calibrated additive reward bonus. Notably, while many papers take advantage of the MCTS algorithm explicitly~\citep{hao2023reasoning,zhou2024language,zhang2024rest,luo2025kbqa,guan2025rstar}, there seems to be a lack of language agents utilizing templates derived from the MCTS algorithm itself. Given the success of MCTS in complex domains like Go and chess~\citep{silver2016mastering,schrittwieser2020mastering}, future work may find an MCTS-inspired agent template useful.

\subsubsection{Divide and Conquer}

Divide and Conquer is an important strategy in computer science and underlies many foundational algorithms \citep{smith1985design, dasgupta2008algorithms, cormen2022introduction}. The basic idea is to breaking a problem into subproblems, solving the subproblems and aggregating the answers. This process can be recursive, meaning that subproblems can be further decomposed into smaller parts. Classic examples include Merge Sort, Strassen matrix multiplication, and Fast Fourier Transform \citep{dasgupta2008algorithms}. For example, Merge Sort breaks sorting a list down to sorting two elements and then merging all the way up, yielding an $\mc{O}(n\log n)$ solution.

For language agents, Divide and Conquer typically means breaking a given problem down to smaller pieces that are solved by the same or different agents  before aggregating the answers. A paradigmatic prompting-based approach following this strategy is Least-to-Most prompting \citep{zhou2022least}---it uses an LLM to explicitly decompose a complex problem into a sequence of simpler subproblems that build upon one another, and then uses the same LLM iteratively solves each subproblem with solutions to earlier subproblems appended to context. For example, an LLM can solve the simple problem ``Alisa has 5 apples. Ben has 2 more apples than Alisa. How many apples do they have together?'' by solving the subproblems ``How many apples does Ben have?'' and ``How many apples do they have together?''. This approach can be formulated as the template $\mc{V} = \{\Lambda_{\mathrm{decomposer}}, \Lambda_{\mathrm{solver}}\}$. Similarly, Decomposed Prompting \citep{khot2022decomposed} introduces a modular framework where subtasks can be recursively decomposed and some subtasks can be accomplished via tool use such as a symbolic retriever. Explicit merge steps are applied here, so $\mc{V} = \{\Lambda_{\mathrm{decomposer}}, \Lambda_{\mathrm{solver}}, \Lambda_{\mathrm{aggregator}}\}$ represents the template. These approaches can improve performance over symbolic reasoning and question answering domains, especially comparing to chain-of-thought prompting. Given the foundational status of the Divide-and-Conquer strategy in algorithm design, it is perhaps no surprise that there is a body of work that applies it to domains that include code generation \citep{zelikman2023parsel} image generation \citep{wang2024divide}, and virtual environments \citep{prasad2024adapt}. 

Beyond problem decomposition with a single underlying LLM, other work has explored using specialized models or agents for different subproblem. HuggingGPT \citep{shen2023hugginggpt} employs a controller LLM that decomposes a user request into subtasks, assigns each to an AI model (e.g., vision, speech) on HugginFace according to their description, and summarize the response accordingly. This allows the system work with multi-modal tasks and problems effectively. Divide-and-conquer agent templates have demonstrated consistent improvements over single-shot approaches. However, many of the Divide and Conquer approaches were proposed before the creation of modern reasoning models \citep{openai2024o1, guo2025deepseek} and the maturation of long-running agents that operate in a multi-turn fashion \citep{yao2023react, wang2024executable, feng2025retool, jin2025search}. This is partly because it has been empirically demonstrated that reasoning models and reasoning-enabled agents can often decompose and solve subtasks on their own \citep[e.g.,][]{gandhi2025cognitive, yang2024swe}. A research frontier is thus how to apply explicit Divide and Conquer templates to the new generation of models for finishing tasks more robustly and solving longer-horizon problems.

\subsection{Templates from RL Algorithms}

\subsubsection{Policy Iteration}

Many of the problems language agents are designed to solve can be expressed as sequential decision-making problem formulated as a finite-horizon, episodic Markov Decision Process (MDP)~\citep{bellman1957markovian,Puterman94} where the state-action space is entirely represented in natural language and all rewards are bounded. Consequently, reinforcement learning algorithms provide a rich source of agent templates.

To the best of our knowledge, one of the earliest applications of an agent template derived from a RL algorithm came in the form of In-Context Policy Iteration (ICPI)~\citep{brooks2023large}. Recall that classic policy iteration (PI)~\citep{howard1960dynamic} leverages dynamic programming~\citep{bertsekas2012dynamic} and assumes access to the true MDP reward function and transition function. PI then proceeds iteratively, beginning with some initial policy $\pi: \mc{S} \ra \mc{A}$ and alternating between two steps. First, there is policy evaluation to compute the action-value function induced by the current policy $Q^{\pi}(s,a)$.
Second, there is greedy policy improvement such that the next policy $\pi'$ is defined as $\pi'(s) = \argmax\limits_{a \in \mc{A}} Q^\pi(s,a)$. Bounded rewards imply the existence of at least one deterministic optimal policy $\pi^\star(s) = \argmax\limits_{a \in \mc{A}} Q^\star(s,a)$ and, for the tabular MDPs PI was originally designed for, a finite state-action space guarantees a finite number of policies to iterate over for achieving global convergence.

In order to implement PI with LLMs, \citet{brooks2023large} derived an agent template consisting of three LLMs, $\mc{V} = \{\Lambda_{\mathrm{policy}}, \Lambda_{\mathrm{transition}}, \Lambda_{\mathrm{reward}}\}$. Relaxing the assumption that the language agent has perfect access to the true environment, ICPI complements the policy LLM, $\Lambda_{\mathrm{policy}}$, needed for PI with two additional LLMs, $\Lambda_{\mathrm{transition}}$ and $ \Lambda_{\mathrm{reward}}$, for approximating the underlying MDP transition and reward function, respectively. The edge structure for the agent template first uses the history of interactions thus far to estimate the recent $\Lambda_{\mathrm{policy}}$ for subsequent policy evaluation. Concretely, the $H$ Bellman backups needed for policy evaluation are ``unrolled'' using the requisite copies of $\mc{V}$. As its name suggests, ICPI uses the full history to prompt all LLMs via in-context learning (ICL)~\citep{brown2020language} and approximate either the MDP model or the most recently executed policy based on collected trajectories. Once policy evaluation is complete, the output is an action greedily chosen based on the collective LLM estimate of $Q^\pi(s,\cdot)$.

ICPI was evaluated across six different MDPs with small state-action spaces and found to perform as well (if not better) than classic tabular $Q$-learning~\citep{watkins1992q}, a fixed rule-based policy searching over all past trajectories, and a random policy performing policy evaluation but not selecting actions via greedy policy improvement. %

\subsubsection{Posterior Sampling for RL}
\label{sec:psrl}

One tractable, statistically-efficient approach to addressing the exploration challenge in RL is to proceed in a Bayesian fashion and sample from the (approximate) posterior distribution over the true underlying MDP. The Bayesian RL setting acknowledges that the environment reward function and transition function are unknown to the decision-maker and, therefore, constitute random variables in the mind of the agent~\citep{ghavamzadeh2015bayesian}. Beginning with some initial, well-specified prior distribution over MDPs, Bayesian RL methods aim to synthesize the Bayes-optimal policy that always strikes the best trade-off between exploration and exploitation. Unfortunately, save for a few tractable cases~\citep{gittins1974dynamic,gittins1979bandit,arumugam2022planning}, the corresponding Bayes-Adaptive MDP~\citep{bellman1959adaptive} that encapsulates this Bayesian sequential decision-making problem is intractable to solve exactly~\citep{duff2002optimal}. To approximately engage with the Bayesian RL problem in a computationally-tractable manner, the RL literature has converged upon posterior-sampling methods that incrementally update posterior beliefs over the underlying MDP. 

One such algorithm is known as Posterior Sampling for Reinforcement Learning~\citep[PSRL;][]{strens2000bayesian}. At each episode, PSRL draws one sample from the current posterior distribution over the true MDP as a statistically-plausible hypothesis for the unknown environment. PSRL employs Thompson Sampling~\citep{thompson1933likelihood,russo2018tutorial} as the mechanism for driving exploration by acting optimally with respect to the posterior sample as if it reflects reality. Given a fully-specified reward function and transition function, any planning algorithm can be used to obtain the optimal policy for this posterior sample, which is then executed for the duration of the episode in the actual environment. The resulting trajectory of experience is a sequence of ground-truth observations from the true reward and transition functions used to perform a posterior update before the next episode. Despite a lack of empirical support outside of tabular MDPs~\citep{osband2013more,osband2017posterior}, PSRL admits elegant theoretical guarantees under various performance criteria (both frequentist as well as Bayesian regret upper bounds) and structural assumptions on the environment, along with a variety of algorithmic extensions~\citep{osband2013more,osband2016posterior,osband2017posterior,lu2019information,arumugam2022deciding}.

An agent template for PSRL was introduced by \citet{arumugam2025toward} consisting of three LLMs $\mc{V} = \{\Lambda_{\mathrm{sample}}, \Lambda_{\mathrm{policy}}, \Lambda_{\mathrm{posterior}}\}$. In lieu of the true Bayesian posterior over MDPs, a compromise of statistical rigor in exchange for tractability was made with a verbal ``posterior,'' summarizing what knowledge the agent currently has and what (task-relevant) uncertainty about the environment remains. In each episode, the resulting LLM-based PSRL agent uses the current ``posterior'' to generate one plausible text description of the underlying MDP with $\Lambda_{\mathrm{sample}}$, deploys the optimal policy for the hypothesized MDP via $\Lambda_{\mathrm{policy}}$, and finally updates the ``posterior'' with $\Lambda_{\mathrm{posterior}}$ to reflect new knowledge as well as residual epistemic uncertainty at the end of an episode. Notably, this agent template does not only consume the current state at each timestep but also carries the verbal ``posterior'' as an additional input and output, providing the consistent algorithmic state needed for all the constituent modules in the template.

Just as PSRL is known for gracefully handling a variety of hard-exploration problems, experiments confirm that a language agent designed with the corresponding PSRL template retains the efficient exploration of the original algorithm. Whereas ICPI assumes that the requisite data needed to facilitate accurate model estimation via ICL has already been collected, \citet{arumugam2025toward} present cumulative regret curves confirming  that this agent template can succeed without such an assumption in difficult natural-language tasks like Wordle~\citep{lokshtanov2022wordle,klissarov2025on}.  

\subsubsection{Information-Directed Sampling}
\label{sec:llm_ids}

While PSRL is one example of a statistically-efficient RL algorithm, its dependence on Thompson Sampling as the underlying exploration mechanism leaves room for improvement. As Thompson Sampling only ever proceeds by an agent acting optimally with respect to current posterior beliefs, it never permits the execution of actions which are deliberately sub-optimal solely for the purpose of gaining information. For example, consider a mail-delivery robot tasked with delivering a package to a new building. PSRL, by leveraging Thompson Sampling, would have this agent spend each episode trying to deliver the package to a new, untested office that has some prior probability of being correct. In contrast, an alternative exploration strategy would be to invest an episode visiting the building directory which, while never part of an optimal trajectory, yields all information needed for behaving optimally thereafter.

To remedy this deficiency, \citet{russo2018learning} propose an algorithmic-design principle known as Information-Directed Sampling (IDS) to accommodate incurring some amount of regret when it yields sufficient information gain. Formally, this is achieved by optimizing for a policy in each time period that strikes a particular balance (specifically, the so-called information ratio~\citep{russo2016information}) between expected regret and information gain. As information gain is quantified by the mutual information between observed experience and optimal actions, which is difficult to compute exactly in general, concrete instantiations of IDS proceed by identifying suitable upper bounds to the information ratio that can be minimized instead to synthesize a behavior policy for each time period. Extensions of IDS beyond the multi-armed bandit setting~\citep{lu2023reinforcement} further require some mechanism for handling non-myopic information gain that may only be obtained by perseverating and deliberately incurring regret across multiple time periods in sequence. Assuming that a suitable surrogate to the information ratio is obtainable, practical instantiations of IDS may then leverage convenient facts from convex analysis~\citep{boyd2004convex} to solve the corresponding policy optimization problem.

While computational efforts with IDS have been restricted to multi-armed bandit problems and smaller-scale MDPs~\citep{lu2023reinforcement}, \citet{arumugam2025toward} outline an agent template directly inspired by IDS. Keeping the machinery of the PSRL agent template for updating a verbal ``posterior,'' the IDS agent template uses two additional LLMs $\mc{V} = \{\Lambda_{\mathrm{regret}}, \Lambda_{\mathrm{info\_gain}}, \Lambda_{\mathrm{posterior}}\}$. Instead of selecting actions via posterior sampling, LLMs $\Lambda_{\mathrm{regret}}$ and $\Lambda_{\mathrm{info\_gain}}$ are prompted to numerically score each action in the current state with an estimate of expected regret and (instantaneous) information gain. For a MDP with $K \in \bN$ actions, the $2K$ LLM-generated values are then used to solve the information-ratio optimization problem of \citet{russo2018learning}, resulting in a probability distribution from which the current action is sampled. 

Theoretical results as well as empirical work on IDS suggests that a suitable instantiation should perform just as well as Thompson Sampling, if not better. For a preliminary investigation in a simpler, numerical variant of the Wordle game, \citet{arumugam2025toward} find that language agents implemented with the corresponding PSRL and IDS templates also preserve this ordering. In particular, while correct guesses for unknown numbers in a target code (rather than a target word, as in traditional Wordle) encumber a LLM-based PSRL agent, a LLM-based IDS agent freely tests all unknown digits before proceeding to explore possible combinations among correct digits; notably, this results in a closer approximation to the Bayes-optimal policy than what the LLM-based PSRL agent could achieve. As the empirical study conducted by \citet{arumugam2025toward} for IDS was preliminary and limited to a single domain, much remains to be seen and understood about the effectiveness of the proposed IDS agent template.

\section{Discussion}

The preceding sections demonstrate how agent templates derived from existing cognitive models or AI algorithms---boasting improved interpretability over arbitrary designs---have already begun to percolate in existing language agents. We discuss alternative views below and provide a call to action for further exploring the potential of this approach.  

\subsection{Alternative Views}

Our advocacy for agent templates derived from cognitive models and existing AI algorithms may be met with a number of reservations, a few of which we address directly here. One alternative position is that we should not invest effort in the general framework of agent templates and instead identify a single best template or agent design pattern. While doing so could bear fruit for some suitably-scoped distribution of tasks, the No Free Lunch Theorem~\citep{wolpert1995nofree,wolpert1997nofree} would imply such a strategy would not be fruitful for all tasks; the precise advantage of having a generic framework of various agent templates is the ability to employ a range of inductive biases suitable to the problem(s) at hand. Our work highlights that a long list of successful pairings between problems and inductive biases has been developed through the literature on cognitive science and AI. We can also make an analogy to the landscape of LLMs themselves---instead of searching for a single ``best'' model, the field recognizes that different kinds of models (in terms of size, modality, architecture, and reasoning) can serve different purposes and use cases. We expect agentic system design to follow a similar pattern.

Another perspective takes no issue with agent templates per se but may posit that LLMs represent a fundamental paradigm shift whereby language agents composed of LLMs face tasks which demand novel templates. We note that, if such a shift does exist, it has yet to make itself apparent in the agents of today; for example, all agents examined in this work for sequential decision-making problems are still RL agents facing some (possibly partially-observable) Markov decision process. Fundamentally, language agents are built to fulfill goals and tasks that cognitive science and AI have long identified (e.g., answering questions, solving hard problems, and interacting with humans). Accepting the premise of this position, we argue that it would still be prudent to first appeal to the wealth of existing templates at our disposal from cognitive science and AI. Doing so would be commensurate with mapping out the manifold of agent designs for tasks of interest thereby highlighting the frontier (if it exists) where novel templates are warranted.

Finally, one may eschew templates inspired by cognitive science and AI in favor of those automatically discovered and generated by LLMs or language agents themselves. While this approach has gained some worthwhile traction \cite{hu2024automated, zhang2024aflow}, those settings are still ones in which cognitive models and AI algorithms may still serve effectively as good starting points or ``priors'' for subsequent design evolution and discovery. Moreover, such designs may further allow us to port over prioritization and preference schemes between the cognitive science or AI-inspired templates themselves; that is, if algorithm $A$ is known to be preferred over algorithm $B$ for a particular class of problem, then one may naturally anticipate the same relation to hold between language agents designed with template $A$ versus template $B$ (see Section \ref{sec:llm_ids} for one such example). Thus, a laissez-faire approach to agent design may not only sacrifice interpretability but also a wealth of prior knowledge about the relative performance of competing language agents.
It is also worth noting that it is possible that such an approach would likely converge on certain agent templates that are highly similar to some cognitive models or AI algorithms. We support this claim by analogy to deep learning, where practitioners are faced with an equally (if not more) onerous challenge of designing effective neural network architectures. 

The deep learning literature gradually converged upon certain staple architectures and design patterns composed of elements from the same core set of building blocks (such as convolutional layers, GRUs, LSTMs, residual blocks, and Transformers). While such designs are not guaranteed to be optimal for any problem of interest, they specify a highly-structured manifold in the space of possible architectures with a demonstrable track record of general reliability across a wide range of problems. Even when formulating the problem of searching over neural network architectures in terms of optimization, researchers still operate within the boundaries defined by these atomic units~\citep{zoph2017neural}, often inspired by or related to intuitions from psychology (for example, CNNs and the visual representations in the brain; the concepts of memory and attention).

\subsection{Call to action}
In this position paper, we have argued that cognitive models and AI algorithms provide effective templates for designing LLM-based agents. Growing together as overlapping fields, cognitive science and AI have respectively developed an abundant collection of ingenious and time-tested computational models and algorithms. As we highlight throughout the paper, many of them have been applied to successful language agent design. 

{\bf We call for researchers to investigate the many other models and algorithms in this space.} Examples include hypothesis generation and learning \citep{dasgupta2017hypotheses, rule2024symbolic}, information-theoretic principles in language \citep{zaslavsky2018efficient, gibson2019efficiency}, and evolutionary algorithms \citep{yu2010introduction, pugh2016quality}. Increasingly, use cases for language agents demand more than completing standalone tasks---they may interact and collaborate with human users, groups of users, or other AI agents \citep{collins2024building, wu2025collabllm}. So, it is conceivable that methods developed in fields that study multi-agent systems and collective decision making (such as economics and computational social science) can well serve as useful templates; among those are voting algorithms \citep{conitzer2024social}, mechanism design concepts \citep{duetting2024mechanism}, and game-theoretic models \citep{sun2025game}. Finally, language agents based on rich, interesting templates may in turn lead to novel or more general cognitive models. Most cognitive models have been limited to working within relatively simple experimental conditions, whereas the open-ended nature of language agents can allow cognitive scientists to predict and explain human thoughts and behaviors in wider and more realistic domains.

\section*{Acknowledgements}

This work was supported by funds provided by the National Science Foundation and by DoD OUSD (R \& E) under Cooperative Agreement PHY-2229929 (the NSF AI Institute for Artificial and Natural Intelligence) and by ONR MURI N00014-24-1-2748. We thank Will Cunningham for conversations and thoughtful feedback on the manuscript.

\bibliography{references}
\bibliographystyle{icml2026}

\end{document}